\documentclass[10pt,twocolumn,letterpaper]{article}

\usepackage{cvpr}
\usepackage{times}
\usepackage{epsfig}
\usepackage{graphicx}
\usepackage{amsmath}
\usepackage{amssymb}
\usepackage{multirow}
\usepackage{soul}
\usepackage[table,xcdraw]{xcolor}
\usepackage[pagebackref=true,breaklinks=true,letterpaper=true,colorlinks,bookmarks=false]{hyperref}

\def\cvprPaperID{9774} % *** Enter the CVPR Paper ID here

\ifcvprfinal\fi
\cvprfinalcopy
\begin{document}

%%%%%%%%% TITLE
%\title{OCD: Over-Complete Distribution Generation For Zero-Shot Learning}
\title{Generalized Zero-Shot Learning Via Over-Complete Distribution}

\author{Rohit Keshari$^+$, Richa Singh$^*$, and Mayank Vatsa$^*$\\
$+$ IIIT-Delhi, India, $*$ IIT Jodhpur, India\\
{\tt\small rohitk@iiitd.ac.in, \{richa, mvatsa\}@iitj.ac.in}}
% For a paper whose authors are all at the same institution,
% omit the following lines up until the closing ``}''.
% Additional authors and addresses can be added with ``\and'',
% just like the second author.
% To save space, use either the email address or home page, not both

\maketitle
%\thispagestyle{empty}

%%%%%%%%% ABSTRACT
\begin{abstract}
   A well trained and generalized deep neural network (DNN) should be robust to both seen and unseen classes. However, the performance of most of the existing supervised DNN algorithms degrade for classes which are unseen in the training set. To learn a discriminative classifier which yields good performance in Zero-Shot Learning (ZSL) settings, we propose to generate an Over-Complete Distribution (OCD) using Conditional Variational Autoencoder (CVAE) of both seen and unseen classes. In order to enforce the separability between classes and reduce the class scatter, we propose the use of Online Batch Triplet Loss (OBTL) and Center Loss (CL) on the generated OCD. The effectiveness of the framework is evaluated using both Zero-Shot Learning and Generalized Zero-Shot Learning protocols on three publicly available benchmark databases, SUN, CUB and AWA2. The results show that generating over-complete distributions and enforcing the classifier to learn a transform function from overlapping to non-overlapping distributions can improve the performance on both seen and unseen classes.
   
   %Moreover, in synthetically generated OCD, the number of hard triplets can be increased and therefore, the inter and intra-class distances are optimized.
\end{abstract}

%%%%%%%%% BODY TEXT
\section{Introduction}

Deep Neural Network (DNN) models have exhibited superlative performance in a variety of real-world applications when the models are trained on large datasets. However, small sample size training sets pose a challenge to deep learning models. It has been observed that in such cases, the DNN models tend to overfit, thus leading to poor generalization. Based on the availability of labeled/unlabeled data, multiple learning paradigms such as transfer learning~\cite{caruana1997multitask}, life-long learning~\cite{thrun1996learning}, self-taught learning~\cite{raina2007self}, and one-shot learning~\cite{miller2002learning} have been proposed for better generalization. The problem becomes further challenging when the training dataset does not contain any sample from the classes in the test dataset. Learning in this scenario is known as zero-data learning or Zero-Shot Learning (ZSL)~\cite{larochelle2008zero}. %Since some of the classes are completely unseen, the performance of deep models is generally not very high. 

\begin{figure}[!t]
     \centering
     \includegraphics[width=0.5\textwidth]{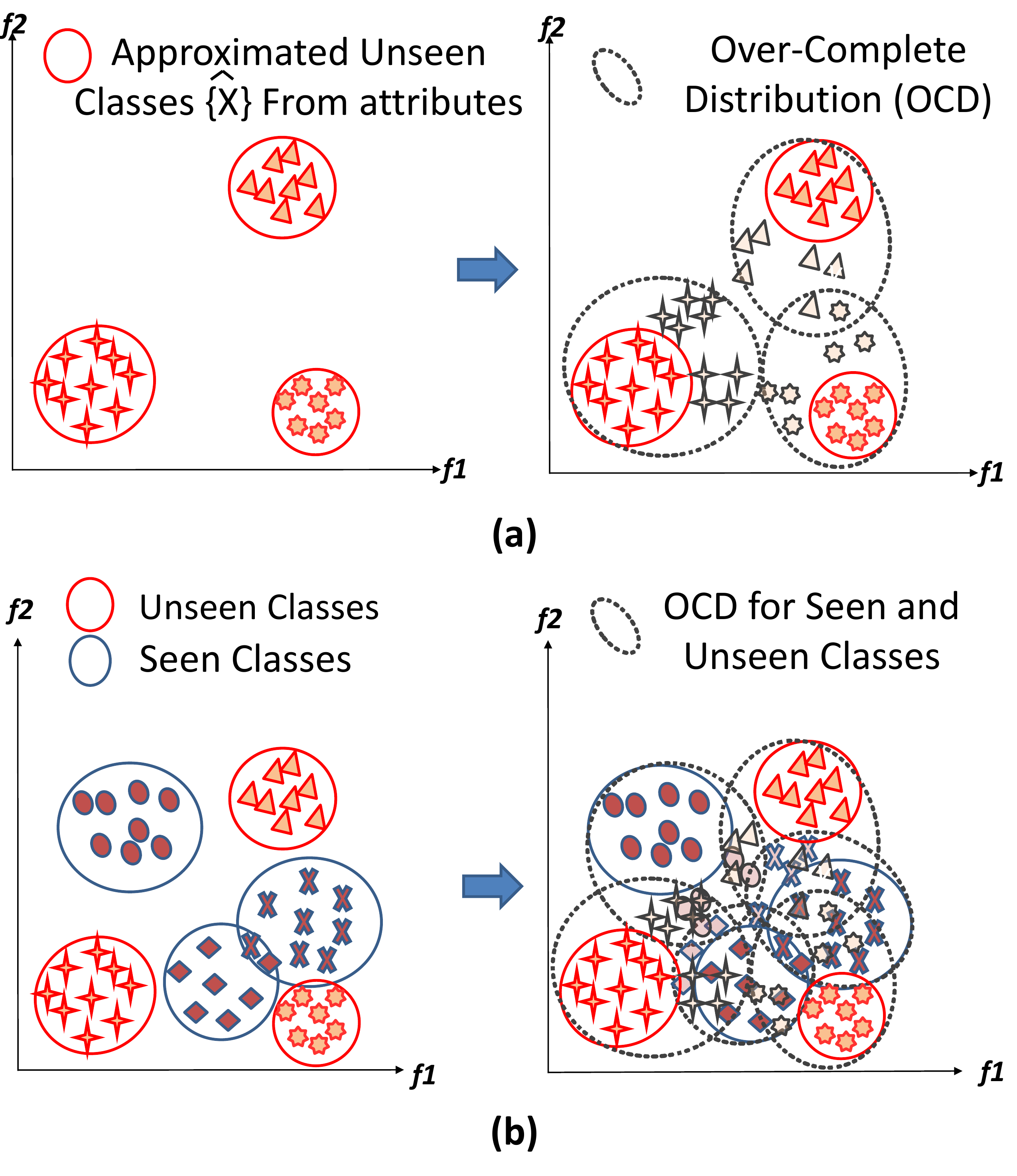}
     \caption{Illustrating seen and unseen 2D distributions before and after the generation of over-complete distribution. $f1$ and $f2$ are two dimensions of the data. (a) Distributions of three approximated unseen classes and generated OCDs for the corresponding classes. (b) Three approximated unseen and seen class distributions, and generated OCDs for the corresponding classes. (Best viewed in colour)}    
     \label{fig:motivation}     
 \end{figure}

To design algorithms for classifying in presence of limited or no training data, researchers have designed two different protocols:  1) conventional Zero-Shot Learning and 2) Generalized Zero-Shot Learning (GZSL) \cite{xian2017zero}. In ZSL problems, dataset is split into two sets with zero intersection in classes and the objective is to maximize the performance on unseen classes. In GZSL, the test dataset contains both unseen and seen classes, and it is required to maximize the performance across both sets of classes. To address the challenges of ZSL and GZSL, researchers have proposed to generate samples pertaining to unseen distribution synthetically~\cite{gao2018joint},~\cite{liu2018generalized},~\cite{kumar2018generalized},~\cite{zhang2019adversarial}. The next section summarizes the research efforts present in the literature. However, the generated unseen classes fail to mimic the real unseen distribution, which is expected to have hard samples. Hence, utilizing synthetically generated classes for training the discriminative classifier does not necessarily improve the performance. 

One of the primary reasons for performance degradation in ZSL is that the testing set contains \textit{hard} samples which are closer to another class and the decision boundary is not optimized for such instances present in the test set. Therefore, it is our assertion that generating hard samples and approximating unseen classes may lead the network to reduce the bias. In this research, we propose the concept of Over-Complete Distribution (OCD).  The objective of over-complete distributions is to generate challenging samples that are closer to other classes, which consequently helps in increasing the generalizability of the network with respect to the unseen classes. Secondly, as shown in Figure~\ref{fig:motivation}, we propose to incorporate Online Batch Triplet Loss (OBTL) to enforce separability between classes and Center Loss (CL) to reduce the spread within the class. We experimentally demonstrate that synthetically generated over-complete distribution allows the classifier to learn a feature space where the separability of seen/unseen classes can be efficiently improved.

\section{Related Work}

The literature in this domain is segregated in two directions: ZSL and GZSL. In ZSL, Larochelle~\etal~\cite{larochelle2008zero} have proposed to learn a mapping from input space view to the model space view. Similarly, Akata~\etal~\cite{akata2013label} have suggested embedding each class into the attribute vector space, called as Attribute Label Embedding (ALE). Liu~\etal~\cite{liu2018generalized} have proposed a Deep Calibration Network (DCN) for learning the common embedding space between the visual features of an image to the semantic representation of its respective class. A widely used method to handle the ZSL problem is to learn a mapping between seen observation to the attribute vector space. Lampert~\etal~\cite{lampert2014attribute} proposed Direct Attribute Prediction (DAP) where a weighted probabilistic classifier has been trained for each attribute. After learning sample-to-attribute mapping, Bayes rule is used to map attributes to the class label. Xian~\etal~\cite{xian2018zero} proposed a more challenging protocol and demonstrated that existing state-of-the-art (SOTA) algorithms do not perform well. %However, the GZSL task is a more challenging protocol proposed by~\cite{xian2018zero}. They have shown that most state-of-the-art algorithms are unable to perform in the GZSL setting. They have also provided a comprehensive evaluation of multiple algorithms on both ZSL and GZSL protocols on their proposed Animal with Attribute2 (AWA2) dataset. 

% synthetically approximated unseen data distribution can be used. Thus, the task of learning a mapping function from observation to class label can be generalized on seen and unseen classes.

In GZSL, researchers have utilized the generated unseen classes to have representative data in the training set~\cite{long2018zero},~\cite{mishra2018generative}. Verma et al.~\cite{kumar2018generalized} have proposed a generative model based on conditional variational autoencoder. They have shown that the synthetically generated unseen distribution is closely approximated to the real unseen data distribution. On synthetically generated data, they have trained supervised linear SVM and shown state-of-the-art performance on the GZSL protocol. Similarly, Gao~\etal~\cite{gao2018joint} have proposed to synthesize the unseen data by utilizing a joint generative model. They have used CVAE and GAN, and observed that preserving the semantic similarities in the reconstruction phase can improve the performance of the model. 

Zhang~\etal~\cite{zhang2019adversarial} have proposed a hybrid model consisting of conditional Generative Adversarial Network (cGAN) and Random Attribute Selection (RAS) for the synthesized data generation. They have trained the hybrid model while optimizing the reconstruction loss. Zhang~\etal~\cite{zhang2019triple} observed that the performance of conventional zero-shot learning algorithms suffer due to Class-level Over-fitting (CO) when they are evaluated for the GZSL task. To overcome the CO problem, they have utilized the triplet loss, which significantly outperforms the state-of-art methods. In another research direction, Long~\etal~\cite{long2018zero} have proposed Unseen Visual Data Synthesis (UVDS) for generating synthesized classes from semantic attributes information. The authors have also proposed Diffusion Regularization (DR), which helps to reduce redundant correlation in the attribute space. Atzmon~\etal~\cite{atzmon2019adaptive} have proposed adaptive confidence smoothing for GZSL problem. They have utilized three classifiers as seen, unseen and gating experts to improve the model performance. Huang~\etal~\cite{huang2019generative} have proposed generative dual adversarial network for learning a mapping function semantic to visual space. Schonfeld~\etal~\cite{schonfeld2019generalized} have proposed to align the distribution generated from VAE and showed improvement on benchmark databases.

%Similarly, Conditional Variational AutoEncoder (CVAE) and Generative Adversarial Networks (GAN) have been used for the GZSL problem~\cite{gao2018joint},~\cite{kumar2018generalized},~\cite{liu2018generalized},~\cite{zhang2019adversarial}. Therefore, semantic similarities in the reconstruction phase can improve the performance of ZSL and GZSL problem. Other than generating the unseen distribution,~\cite{zhang2019triple} have shown that triplet based verification model provides competitive results on both ZSL and GZSL settings.
%%%%

%In traditional ZSL, a dataset is divided into training and testing sets have zero overlapping classes. 

%%%%

Significant efforts have been made in the direction of generating unseen synthetic distributions for training the model. However, as discussed earlier, there are still challenges to be addressed to improve the performance on ZSL and GZSL problems, such as generalization of the model on the test set and reduce the bias for both seen and unseen classes.    

%learning a discriminative classifier on unseen/seen classes which can transform the overlapped distribution to another space where classes are separable among seen/unseen classes ( has been less explored.  

%Attributes of a sample are utilized to generate synthetic data corresponding to unseen samples. 
%Despite all the advancement in ZSL, algorithms still perform poorly on the GZSL settings. In this research, our proposed method strictly follows the standard protocols explained by Xian~\textit{et al.}~\cite{xian2018zero}. Moreover, we focused on building a robust classifier based on the generation of hard triplets such that the class separability can be imposed on the seen/unseen classes.

%\section{Related Work}
%Recently, zero-shot learning has received a significant amount of interest from the research community. Therefore, considerable improvement and standardized protocols have been made by the researchers. Due to the space limitation, the summarization of all works is not possible. However, to the best of our knowledge, the following paragraphs contain the most recent advancement related to our work. 

\section{Proposed Framework}

\begin{figure*}[!t]
     \centering
     \includegraphics[width=1.0\textwidth]{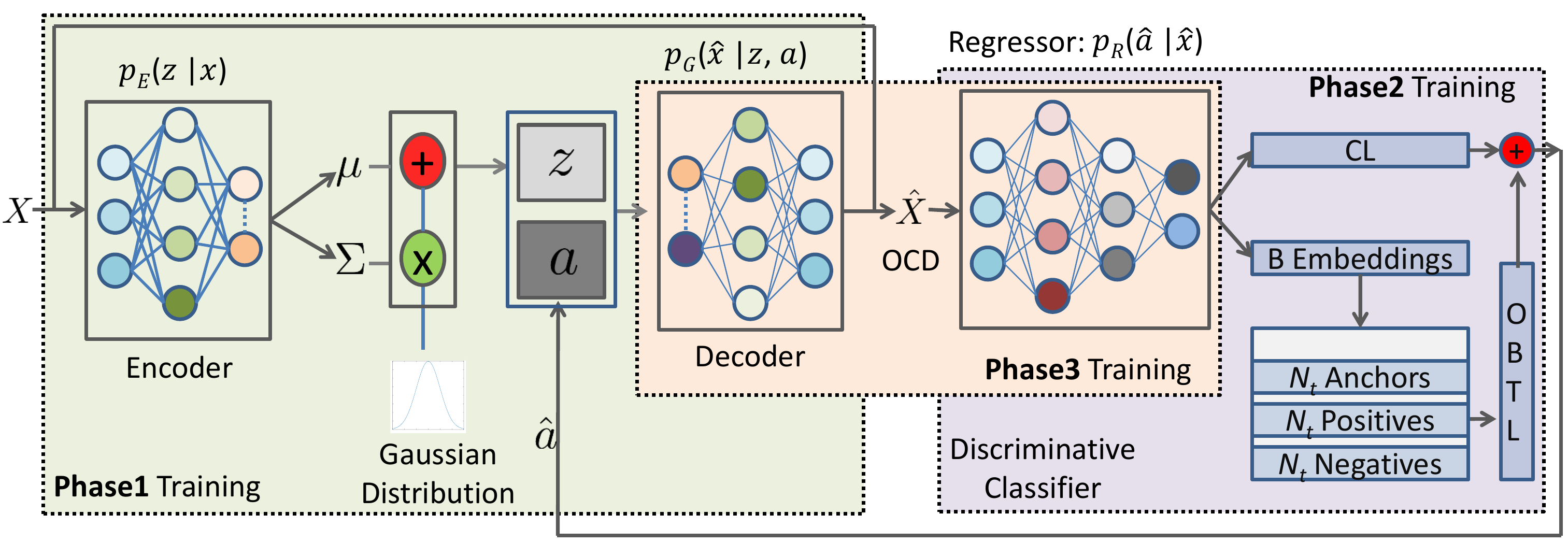}
     \caption{Illustration of the proposed OCD-CVAE framework. The framework uses Conditional Variational AutoEncoder (CVAE) with encoder $p_E(z|x)$ and decoder $p_G(\hat{x}|z,a)$ modules. The output of CVAE is given to the regressor $p_R(\hat{a}|\hat{x})$ where regressor maps the generated samples to its respective attributes. To generate the unseen synthetic data, attributes of unseen samples and randomly sampled $z$ are provided to the trained decoder.}    
     \label{fig:block}     
 \end{figure*}

%The proposed approach is inspired by the research in synthetic generation of unseen data~\cite{gao2018joint},~\cite{kumar2018generalized},~\cite{liu2018generalized},~\cite{zhang2019adversarial} which helps in learning the class separability by the regressor. The complete pipeline of the proposed algorithm is shown in Figure~\ref{fig:block}. 

Figure~\ref{fig:block} demonstrate the steps involved in the proposed framework. For a given input $x$ with associated attribute $a$ and latent variable $z$, there are three modules in the proposed pipeline: (i) an encoder ($p_E(z|x)$) to compute the latent variables $z$ on given $x$, (ii) a decoder ($p_G(\hat{x}|z,a)$) to generate samples $\hat{x}$ on given $z$ and attribute $a$, and (iii) a regressor ($p_R(\hat{a}|\hat{x})$) to map $\hat{x}$ to their predicted attribute $\hat{a}$. The combined encoder and decoder modules is called as CVAE, which has been conditioned on attribute $a$. The regressor module is trained with the OBTL and CL losses to optimize the interclass and intraclass distances. This section presents the details of each of the modules followed by the training process and the implementation details.

%\textcolor{red}{The proposed framework can be explained based on the output of the modules: 1) output of the CVAE which can be utilized for the generation of OCD, and 2) output of the regressor where interclass/intraclass distance can be optimized while utilizing OBTL and CL losses. Moreover, training the proposed framework has been divided into three phases. In the first and second phase of training, CVAE and regressor are trained and treated as pre-trained models for the third phase of training.}
  
 %\subsection{Preliminaries}
%The proposed OBTL is a modified version of triplet loss. Along with triplet loss, center loss has been widely utilized in the metric learning literature. Therefore, we first briefly introduce both loss functions. \\
%OBTL and CL have been utilized for the optimization of interclass/intraclass distance in the proposed framework. We have experimentally shown that optimizing these loss functions improve the proposed framework performance. 

\begin{figure}[!t]
     \centering
     \includegraphics[width=0.45\textwidth]{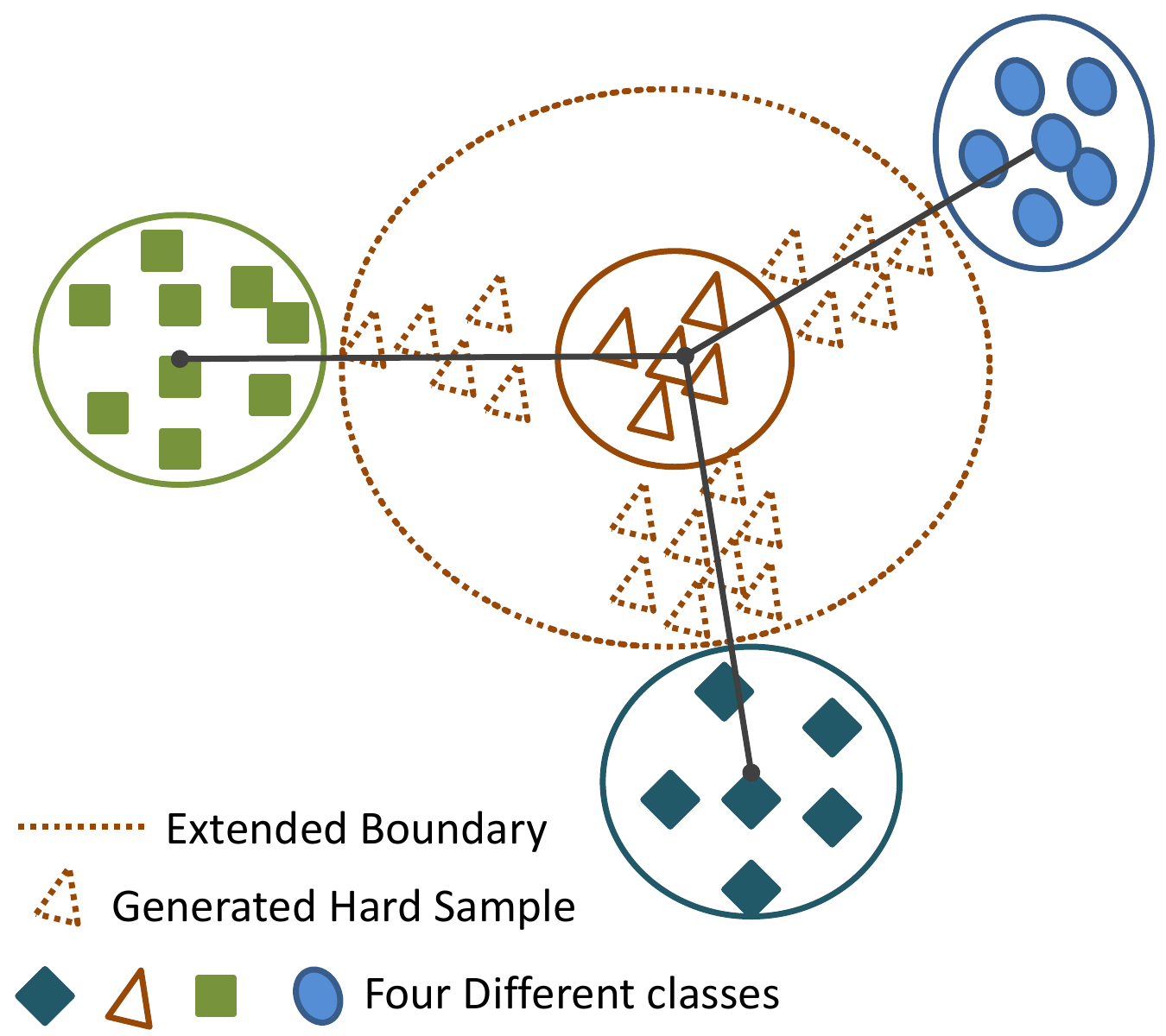}
     \caption{Illustration of over-complete distribution generation while generating hard samples between two classes. The boundary of the distribution would be decided from equations~\ref{eq:ocd1} and~\ref{eq:ocd2}. Since $\mu_{OC}$ is the average between one class to other competing classes, the boundary would be extended based on the new obtained $\mu_{OC}$.}    
     \label{fig:hard}     
 \end{figure}
 
\subsection{Over-Complete Distribution (OCD)}

% \subsection{Regressor Loss for Inter and Intra class Distance Optimization}

%To generalize the network on unseen distribution for ZSL and GZSL tasks, the biasing problem towards seen classes has to be minimized. Therefore, the decision boundary of the regressor can be generalized on the seen/unseen class distribution. Thus, the regressor should robust enough to transform the real unseen class distribution to another space where classes are separated. Hence, generating a hard sample is required, which further can be utilized to train the classifier while enforcing the class separability. 
% There are multiple methods for generating hard samples~\cite{yuan2017hard},~\cite{zhao2018adversarial},~\cite{zheng2019hardness} have been proposed. However, synthetically creating negative samples on unseen classes has been less explored. 

%Therefore, we have termed it over-complete and the real distribution of a class might not contain hard samples for each class of the database. 

The primary task of the decoder (shown in Figure~\ref{fig:block}) is to generate or approximate a distribution which is closer to the real unseen data. As shown in Figure~\ref{fig:hard}, creating the OCD for a class involves generating all the possible hard samples which are closer to other class-distributions. 
Since simulating the behaviour of real unseen distribution is a challenging problem, we first propose to generate OCD for a class and visually show that the generated OCD simulates the behaviour of the real unseen distribution. Using the given distribution, OCD is generated by mixing a finite number of multiple Gaussian distributions~\cite{richardson1997bayesian} while shifting the mean towards other classes. If the distribution is not known (in case of unseen classes), the distribution of the class can be approximated by using generative models. The parameters of approximated distribution from the variational inference of a class are represented by $\mu$, $\sigma$ and the over-complete distribution is represented via $\mu_{OC}$, $\sigma_{OC}$, where $\sigma_{OC}>\sigma$. 

Let $\hat{X}$, and $\hat{X}_{OC}$ be the approximated unseen distribution and over-complete distribution, respectively.

% \begin{align}
%     \hat{X}&= p_G(x|\mathcal{N}(\mu_{HP},\sigma_{HP}),a)\; and\; \hat{Z}=p_E(z|\hat{x}) \;,where\;\hat{x}\sim \hat{X},~\mu_{z|\hat{X}},~\sigma_{z|\hat{X}} \label{eq:ocd1} \\
%     \hat{X}_{OC}&= p_G(x|\mathcal{N}(\mu_{OC},\sigma^{'}_{HP}),a),\; where\; \mu_{OC}= \frac{\mu_{z|\hat{X}}+\mu^{'}_{z|\hat{X}}}{2},\; and\; \mu^{'}_{z|\hat{X}}= \mu_{z|\hat{X}}[j] \label{eq:ocd2} 
% \end{align}

\begin{multline}
\hat{X}= p_G(x|\mathcal{N}(\mu_{HP},\sigma_{HP}),a)\; and\; \hat{Z}=p_E(z|\hat{x}) \;,\\ where\;\hat{x}\sim \hat{X},~\mu_{z|\hat{X}},~\sigma_{z|\hat{X}} \label{eq:ocd1}
\end{multline}
    
\begin{multline}    
    \hat{X}_{OC}= p_G(x|\mathcal{N}(\mu_{OC},\sigma^{'}_{HP}),a),\\ \mu_{OC}= \frac{\mu_{z|\hat{X}}+\mu^{'}_{z|\hat{X}}}{2},\; \mu^{'}_{z|\hat{X}}= \mu_{z|\hat{X}}[j] 
    \label{eq:ocd2}
\end{multline}

\begin{figure*}[!ht]
     \centering
     \includegraphics[width=1.0\textwidth]{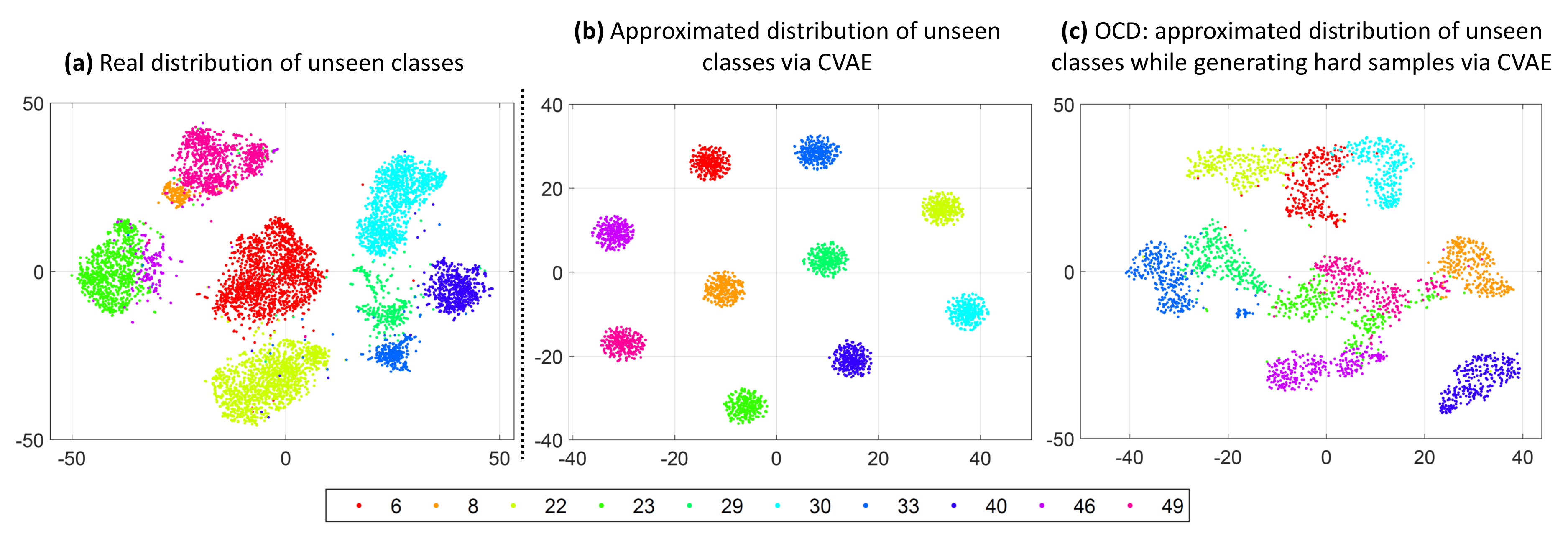}
     \caption{Illustrating synthetically generated distributions for unseen classes of the AWA2 database. (a) Real unseen distribution of the AWA2 database. Different colours represent different classes of the AWA2 database with the PS protocol, (b) Approximated distribution of unseen classes via CVAE, and (c) Approximated over-complete distribution of unseen classes via CVAE. From the distribution plot, it can be observed that (c) is a closer representation of the real unseen class distribution shown in (a). (Best viewed in colour)}    
     \label{fig:dist_plot}     
 \end{figure*}

Equations~\ref{eq:ocd1} and~\ref{eq:ocd2} represent the process of generating the over-complete distribution. Here, $p_G(.)$ is a generator module of the pipeline. $\mu_{HP}$ and $\sigma_{HP}$ are the hyper-parameters for normal distribution. $\mu_{z|\hat{X}}$ and $\sigma_{z|\hat{X}}$ are mean and standard deviation obtained while encoding the data $\hat{X}$ into the latent space, $z$. In Equation~\ref{eq:ocd2}, $\sigma^{'}_{HP}$ is a hyper-parameter and $j$ is a randomly sampled index variable for shuffling of the parameter $\mu_{z|\hat{X}}$. In both the equations, $\mathcal{N}(.)$ is a Gaussian distribution generator.

In the first part of Equation~\ref{eq:ocd1}, distribution of unseen classes, $\hat{X}$, is generated by randomly sampling $z\sim \mathcal{N}(\mu_{HP},\sigma_{HP})$ with $\mu_{HP}$, $\sigma_{HP}$ as the parameters of the distribution and unseen class attributes $a$. In the second part of Equation~\ref{eq:ocd1}, $\mu_{z|\hat{X}}$ and $\sigma_{z|\hat{X}}$ are estimated by using the encoder module $p_E(.)$. The first part of Equation~\ref{eq:ocd2} represents the generation of over-complete distribution $\hat{X}_{OC}$ in which latent variable $z\sim \mathcal{N}(\mu_{OC},\sigma^{'}_{HP})$ is randomly sampled from the Gaussian distribution, where, mean of the distribution $\mu_{OC}$ is estimated by the average of the current and each competing class. For example, on a given batch of $\mu$, $j$ is an index variable ranging from ${1,..., batch\;size}$, and it is randomly sampled without repetition.

%For ZSL, on the trained CVAE, distribution of unseen classes, $\hat{X}$, is generated by randomly sampling $z$ from Gaussian distribution of parameters $\mu_{HP}$, $\sigma_{HP}$ and unseen class attributes (first part of Equation~\ref{eq:ocd1}). After generating $\hat{X}$, $\mu_{z|\hat{X}}$ and $\sigma_{z|\hat{X}}$ are estimated by using the encoder module of the CVAE (the second part of Equation~\ref{eq:ocd1}). 

In our approach, the decoder/generator $P_G(x|z, a)$ is conditioned on attributes and used in Equations~\ref{eq:ocd1} and~\ref{eq:ocd2}. In ZSL problems, it is assumed that attributes are a good representation of a class and separated in the attribute space. Within a class, if a sample is away from the centre of the distribution, then it can be considered as a hard sample. However, attributes of the sample should be the same as the attributes of a class. Therefore, while generating the OCD, attributes of a class are kept unchanged. On the other hand, the latent variable $z$ has been changed based on the mean parameter of other classes.\\

\noindent\textbf{Visualization of the Distributions: } Figure~\ref{fig:dist_plot}(b) shows the unseen distribution predicted/generated via CVAE, and the classes are well-separated. However, as shown in Figure~\ref{fig:dist_plot}(a), the real distribution of unseen classes are closer to other classes and some of them are overlapping. If the generated distribution fails to mimic the behaviour of real distribution, then the utility of such distributions is limited in training. Usually, the discriminative classifier trained on such distribution performs poorly on unseen classes. On the other hand, learning class separability by maximizing inter-class distance and minimizing intra-class distance might provide a viable solution when the behaviour of the training set is close to the test set. In the case of ZSL, distribution is unknown and approximating unknown distribution where latent variables are sampled from Gaussian distribution can lead to blind-spots in the feature space. As observed in Figure~\ref{fig:dist_plot}(b), blind-spot is the place where samples are not present and training a classifier with such data would not ensure the model to learn transformations which are effective for separating real unseen distributions. Figure~\ref{fig:dist_plot}(c) illustrates the OCD, which is an approximated distribution of unseen classes while generating hard samples via CVAE. Experimentally we have shown that training using such distribution can improve the classification performance.

\subsection{Proposed OCD-CVAE Framework Training}

As shown in Figure~\ref{fig:block}, we propose to train the OCD-CVAE framework in three phases. In the first phase, CVAE loss ($\mathcal{L}_{CVAE}$) is optimized. In the second phase, OBTL loss along with center loss, i.e., $\mathcal{L}_{OBTL}+\mathcal{L}_{CL}$, is minimized. The trained model is then utilized as a pre-trained model for the third phase, where we propose training the regressor on the generated OCD while minimizing the Online Batch Triplet Loss (OBTL) and CL losses. In this section, we first discuss the loss functions, followed by the details of the three training phases.

%The embedding space created from the output of the regressor, increases the inter-class distance and reduces the intra-class distance.

\subsubsection{Loss Functions}

\noindent \textbf{Online Batch Triplet Loss to Maximize Inter Class Distance:} The triplet loss has been widely used in the literature to increase the inter-class distance ($D_{inter}$) and decrease the intra-class distance ($D_{intra}$). Mathematically, triplet loss can be represented as:

\begin{equation}
\label{eq:triplet}
    \mathcal{L}_t(f^a,f^p,f^n)=\sum_{i=1}^N \Big[||f^a_i-f^p_i||_2^2-||f^a_i-f^n_i||_2^2+\alpha \Big]_+
\end{equation} 

\noindent where, $f$ represents the embedded feature vector, $\mathcal{L}_t$ is the triplet loss and triplet $(f^a,f^p,f^n)$ is a 3-tuple of anchor, positive, and negative, respectively. $\alpha$ represents the margin to control the distance between the positive and negative pairs. $\big[\big]_+=max(0,.)$ represents the hinge loss function.  

If $D_{inter}<D_{intra}$ then triplet $<A, P, N>$ is considered to be a hard triplet. From Equation~\ref{eq:triplet}, it can be observed that $\mathcal{L}_t(f^a,f^p,f^n)>0$ only if $||f^a-f^p||_2^2+\alpha > ||f^a-f^n||_2^2$. Therefore, hard triplet mining is an essential step to minimize triplet loss. 

%\footnote{Inter-class distance $D_{inter}$ (distance between anchor A and negative sample N) should be maximized and intra-class distance $D_{intra}$ (distance between anchor A and positive sample P) should be minimized.

As shown in Figure~\ref{fig:hard}, the generated hard samples between two classes lead to generating the over-complete distribution for a class. The approximated OCD is then utilized for training a discriminative classifier for triplet loss minimization. Selecting $\mathcal{N}_t$ hard triplets in offline mode requires processing all the generated triplets in a single epoch which is very challenging in real-world ZSL settings. 
Therefore, we propose the Online Batch Triplet Loss, which is inspired by the online triplet loss~\cite{Amos2016OpenFace}. Generating triplets in a batch-wise manner reduces the search space to find hard negative samples and the total training time of the deep model\footnote{For instance, we have 20 samples per class from 10 classes in the dataset. Selecting every combination of 2 images from each class for the anchor and positive images and then selecting a hard-negative from the remaining images gives $10\times(C^{20}_2)$ = 1900 triplets. Despite 200 unique samples, it requires 19 forward and backwards passes to process 100 triplets at a time. In OBTL, these embeddings are mapped to 1900 triplets that are passed to the triplet loss function, and then the derivative is mapped back through to the original sample for the backwards network pass - all with a single forward and single backward pass.}. 

The proposed $\mathcal{L}_{OBTL}$ minimizes the generated triplets for every batch while training the model. $\mathcal{L}_{OBTL}$ is optimized in a manner similar to $\mathcal{L}_t$ as defined in Equation~\ref{eq:triplet}. It is our assertion that synthetically generating hard negatives can improve the learning of the deep model. \\ %On the other hand, applicability on the data would be changed in case of $\mathcal{L}_{OBTL}$. 

\noindent \textbf{Center Loss:} Mapping a sample to their attributes has been used to find a solution for the ZSL problem. In order to learn the mapping of different samples to the attribute of a class, the standard deviation of a class distribution in the attribute space should be minimum. Therefore, center loss~\cite{wen2016discriminative}, along with regressor loss~\cite{kumar2018generalized} has been utilized to minimize the deviation from the center. 

%$p_R(a|x)$ 
As shown in Figure~\ref{fig:block}, the regressor maps the approximated $x$ to the associated attribute $a$. Since hard samples increase the standard deviation, it is important to minimize the centre loss for the over-complete distribution. Therefore, the discriminative classifier is trained with center loss $\mathcal{L}_{CL}$:

%While predicting the distribution of the unseen classes, the distribution has been predicted based on the sampling from a random distribution. To aid in triplet loss learning, hard negative has been generated. Therefore, synthetically generated data has a higher standard deviation.

\begin{equation}
    %\mathcal{L}_{CL}=\frac{1}{2}\sum_{c=1}^{S+U}||x_c-CT_{c}||_2^2
    \mathcal{L}_{CL}=\frac{1}{2}\sum_{c=1}^{S+U}||x_c-x^{CT}_c||_2^2
\end{equation}

\noindent where, $x_c$ represents a sample from class $c$ and $x^{CT}_c$ is the learned center of class $c$.

\subsubsection{Learning Phase of the Proposed Model}
As shown in Figure~\ref{fig:block}, the learning phase of the proposed framework can be divided into three phases. In the \textit{first phase}, encoder followed by decoder (CVAE) is trained using KL-divergence and conditional marginal likelihood. In the \textit{second phase}, regressor/classifier is trained using the proposed OBTL along with CL. In the \textit{third phase}, the decoder/generator and the regressor have been trained while minimizing the OBTL, CL, and discriminator driven losses~\cite{kumar2018generalized}. 

Let the training set contain `$S$' seen classes and the testing set contain `$U$' unseen classes. Their respective class attributes are represented as $\{a_c\}_{c=1}^{S+U}$ where, $a_c\in \mathbb{R}^L$ and $L$ is the length of attributes. Training $\mathcal{D}_S$ and testing $\mathcal{D}_U$ sets can be represented as the triplet of data, attributes, and label $\{X_s,a_s,y_s\}_{s=1}^{S}$ and $\{X_u,a_u,y_u\}_{u=1}^{U}$, respectively. On the aforementioned settings, ZSL algorithm aims to build a classification model on $\mathcal{D}_S$ which can learn a mapping function $f: \mathcal{X}_U \rightarrow \mathcal{Y}_U$ where, $\mathcal{X}_U=\{X_u\}_{u=1}^{U}$ is a set of unseen samples and $\mathcal{Y}_U=\{y_u\}_{u=1}^{U}$ is the corresponding class set~\cite{akata2013label,lampert2009learning}. \\ 

\noindent\textbf{First phase of training:} In the first phase of training, CVAE is trained on $\mathcal{D}_S$, where the input sample is $x_i$ for the encoder which encodes the latent variable $z_i$. Encoded variable is appended with the attributes $a_i$ of the corresponding sample. The appended latent variable $[z_i, a_i ]$ is then provided to the generator module that generates the outputs $\hat{x}_i$ for a particular distribution which is close to the input provided to the encoder module. Trained CVAE allows the decoder to generate synthetic data on given attributes $a$.  The CVAE loss ($\mathcal{L}_{CVAE}$) can be defined as:

\begin{multline}
    \mathcal{L}_{CVAE}=-\mathbb{E}_{p_{E(z|x)},p_{(a|x)}}[log_{p_G}(\hat{x}|z,a)]+\\ KL(p_E(z|x)||p(z))
\end{multline}

\noindent where, $-\mathbb{E}_{p_{E(z|x)},p_{(a|x)}}[log_{p_G}(\hat{x}|z,a)]$ is the conditional marginal likelihood and $KL(p_E(z|x)||p(z))$ is the KL-divergence. Inspired from Hu et al.~\cite{hu2017toward}, the joint distribution over the latent code $[z, a]$ is factorized into two components $p_E(z|x)$ and $p_R(\hat{a}|x)$ as a disentangled representation. 

\vspace{6pt}
\noindent\textbf{Second phase of training:} In the second phase of training, regressor is trained on $\mathcal{D}_S$ while minimizing the two losses \\
\begin{equation}
\min_{\theta_R}\mathcal{L}_{OBTL}+\mathcal{L}_{CL}  
\end{equation}
The regressor is trained to improve the mapping of generated synthetic data to the corresponding attribute.\\

\noindent\textbf{Third phase of training:} In the third phase of the training, $D_s$ and approximated OCD have been utilized. From the first phase, we obtain $\theta_G$ (the generator parameter) which is used in the third phase of training. In the third phase, loss $\mathcal{L}_c(\theta_G)$ is based on the discriminator prediction and $\mathcal{L}_{Reg}(\theta_G)$ is used as a regularizer.

\begin{equation}
\begin{split}
    \mathcal{L}_c(\theta_G)=-\mathbb{E}_{p_G{(\hat{x}|z,a)}p(z)p(a)}[log\;{p_R}(a|\hat{x})] \\
    \mathcal{L}_{Reg}(\theta_G)=-\mathbb{E}_{p(z)p(a)}[log\;{p_G}(\hat{x}|z,a)]
\end{split}
\end{equation}

\noindent This regularization is used to ensure that the generated OCD produces class-specific samples, even if $z$ is randomly sampled from $p(z)$. The complete objective function of the third phase can be represented using the equation below where, $\lambda_c$, and $\lambda_{reg}$ are the hyper-parameters.

%, and $\mathcal{L}_E(\theta_G)$ has been defined as $-\mathbb{E}_{\hat{x}\sim p_G(\hat{x}|z,a)}KL[(p_E(z|\hat{x})||q(z))]$.

\begin{equation}
    \min_{\theta_G,\theta_R} \left ( \lambda_c . \mathcal{L}_c+\lambda_{reg}.\mathcal{L}_{Reg}+ \mathcal{L}_{OBTL}+\mathcal{L}_{CL}\right )
\end{equation}

\subsection{Implementation Details}

Experiments are performed on a 1080Ti GPU using Tensorflow-1.12.0~\cite{abadi2015tensorflow}. Hyper-parameters for CVAE learning: $\lambda_c=0.1$, $\lambda_R=0.1$, $\lambda_{reg}=0.1$, and $batch\; size = 256$. To generate hard samples, the value of hyper-parameters $\mu_{HP}$, $\sigma_{HP}$ and $\sigma^{'}_{HP}$ in Equations~\ref{eq:ocd1} and~\ref{eq:ocd2} are $0$, $0.12$, and $0.5$, respectively. In our experiments, the size of $\mu$ is $256 \times 100$ and row-wise shuffling is performed. %For reproducibility, we will release the model files of the proposed algorithm.

%The value of hyper-parameters such as epoch, learning rate, the batch size is kept as $10^-5$, $[10^{-1},...,10^{-5}]$, 64 respectively for all the experis. Learning rate is started with $10^{-1}$ and keep reducing by the factor $10$ at every 100 epoch. For CIFAR10/CIFAR100/SVHN, $pw$ is set as 3 and stride as $2$. For Tiny ImageNet, $pw$ is 6 and stride is $2$. Empirically, we have found that four softmax layers are an optimal number of the classifier for the deep model, hence according to the database pooling window $pw$ and stride are adjusted. For the reproducibility, we will release the source code of the proposed model.

\section{Experimental Results and Analysis}

%The proposed algorithm is evaluated on both ZSL and GZSL scenarios on the . 
The proposed framework is evaluated on both ZSL and GZSL settings, and compared with recent state-of-the-art algorithms. This section briefly presents the databases and evaluation protocols, followed by the results and analysis on the AWA2~\cite{lampert2014attribute}, CUB~\cite{welinder2010caltech}, and SUN~\cite{patterson2012sun} benchmarking databases.%In both the protocols, the proposed framework improves the existing performance on the three databases mentioned above.  

\subsection{Database Details}
The statistics and protocols of the databases are presented in Table~\ref{tb:db}. All databases have seen/unseen splits as well as attributes of the corresponding classes. The Animals with Attributes2 (AWA2)~\cite{lampert2014attribute} is the extension of the AWA~\cite{lampert2009learning} database containing $37,322$ samples. It has images from $50$ classes and size of attribute vector is $85$, both these are consistent with the AWA database. The $85$ dimensional attributes are manually marked by humans experts. The Caltech UCSD Bird 200 (CUB)~\cite{welinder2010caltech} database contains $11,788$ fine-grained images of $200$ bird species. The size of the attribute vector is 312. The SUN Scene Classification (SUN)~\cite{patterson2012sun} database contains $14,204$ samples of 717 scenes. It has an attribute vector of $102$ length.

\begin{table}[!t]
\begin{center}
\small
\caption{Databases used in the experiments.}
\label{tb:db}
\begin{tabular}{|c|c|c|c|}
\hline
\multicolumn{1}{|c|}{\textbf{Dataset}} & \multicolumn{1}{c|}{\textbf{Seen/Unseen Classes}} & \multicolumn{1}{c|}{\textbf{Images}} & \multicolumn{1}{c|}{\textbf{Attribute-Dim}} \\ \hline
SUN                                    & 645/72                                    & 14340                                 & 102                               \\ \hline
CUB                                    & 150/50                                    & 11788                                 & 312                               \\ \hline
AWA2                                   & 40/10                                     & 37322                                 & 85                                \\ \hline
\end{tabular}
\end{center}
\end{table}

%\begin{table}[!t]
%\centering
%\scriptsize
% \footnotesize
%\caption{Ablative study on 3 datasets with the PS protocol. The reported values are classification accuracy (\%).}
%\label{tb:ablation}
%\begin{tabular}{|c|c|c|c|c|c|}
%\hline
%\textbf{Dataset (PS)} &  \textbf{OBTL} & %\textbf{CL} & \textbf{OCD+OBTL} & \textbf{OCD+CL} & \textbf{OCD+OBTL+CL} \\ \hline
%\textbf{AWA2}         & 65.8         & 65.3        & 70.9             & 66.5            & \textbf{71.3}                \\ \hline
%\textbf{SUN}          & 56.4         & 56.2  %      & 62.0               & 57.6            & \textbf{62.1}                \\ \hline
%\textbf{CUB}          & 54.5         & 53.7        & 60.5             & 56.8            & \textbf{60.9}                \\ \hline
%\end{tabular}
%\end{table}

\subsection{Evaluation Protocol}

The experiments are performed with both Zero-Shot Learning and Generalized Zero-Shot Learning protocols. In ZSL, OCD for unseen classes are generated and utilized to train the proposed OCD+CVAE framework. The results are reported on both standard split (SS) given by Larochelle~\etal~\cite{larochelle2008zero} and proposed split (PS) given by Xian~\etal~\cite{xian2018zero} protocols. Unseen class classification accuracies are reported for both PS and SS protocols. 

For GZSL, the seen classes of the dataset are divided into 80-20 train-test ratio to obtain the two sets: $X^S_{train}$ and $X^S_{test}$. The set $S+U$ is used for the training where $U$ has been synthetically generated by the generator module of the proposed framework. For testing, the model is evaluated on $X^U$ and $X^S_{test}$. In GZSL, as defined in the literature, average class accuracies of protocols $\mathbf{A}$ and $\mathbf{B}$ are reported. Protocol $\mathbf{A}$ is an average per-class classification accuracy on $X^U_{test}$ where, a regressor is trained on $S+U$ classes ($\mathbf{A}:U\rightarrow S+U$). Protocol $\mathbf{B}$ is an average per-class classification accuracy on $X^S_{test}$ where, a regressor is trained for $S+U$ classes ($\mathbf{B}:S\rightarrow S+U$). The above mentioned protocols are predefined for AWA2~\cite{lampert2014attribute}, CUB~\cite{welinder2010caltech}, and SUN~\cite{patterson2012sun} databases and widely used to evaluate ZSL/GZSL algorithms. The proposed model maps a sample to the corresponding attribute.

%$X^S_{train}$, and $X^S_{test}$ sets are obtained by randomly divided the dataset into 80-20 ratio. The reported results are evaluated on $X^U$ and $X^S_{test}$. $S+U$ is a set used for the training where $U$ has been synthetically generated by generator module of the framework.

% \begin{figure}[!t]
%      \centering
%      \includegraphics[width=0.5\textwidth]{figures/zero-shot_db.pdf}
%      \caption{Sample images of the SUN, CUB, and AWA2 databases.}    
%      \label{fig:db}     
%  \end{figure}

%On the AWA2 database, the proposed algorithm shows an improvement of $1.8\%$. On CUB and SUN databases, the proposed algorithm is the second best performing algorithm. Moreover, 

%The proposed framework is performing close to the state-of-art methods.

\begin{table}[!t]
\centering
\footnotesize
\caption{Classification accuracy (\%) of conventional zero-shot learning for standard split (SS) and proposed split (PS)~\cite{xian2018zero}. (Top two performances are in bold)}
\label{tb:zsl}
\begin{tabular}{|l|l|l||l|l||l|l|}
\hline
\multicolumn{1}{|c|}{\multirow{2}{*}{\textbf{Method}}}      & \multicolumn{2}{c||}{\textbf{AWA2}} & \multicolumn{2}{c||}{\textbf{CUB}} & \multicolumn{2}{c|}{\textbf{SUN}} \\ \cline{2-7} 
\multicolumn{1}{|c|}{}                                      & \textbf{SS}     & \textbf{PS}     & \textbf{SS}     & \textbf{PS}     & \textbf{SS}      & \textbf{PS}     \\ \hline\hline
CONSE~\cite{norouzi2013zero}          & 67.9             & 44.5            & 36.7            & 34.3            & 44.2            & 38.8            \\ \hline
SSE~\cite{zhang2016learning}          & 67.5             & 61.0           & 43.7            & 43.9            & 54.5            & 51.5              \\ \hline
LATEM~\cite{xian2016latent}           & 68.7             & 55.8            & 49.4            & 49.3            & 56.9            & 55.3            \\ \hline
ALE~\cite{akata2013label}             &  80.3             & 62.5           & 53.2            & 54.9            & 59.1            & 58.1            \\ \hline
DEVISE~\cite{frome2013devise}         & 68.6             & 59.7            & 53.2            & 52.0              & 57.5            & 56.5            \\ \hline
SJE~\cite{akata2015evaluation}        & 69.5             & 61.9            & 55.3            & 53.9            & 57.1            & 53.7            \\ \hline
ESZSL~\cite{romera2015embarrassingly} & 75.6             & 58.6            & 55.1            & 53.9            & 57.3            & 54.5            \\ \hline
SYNC~\cite{changpinyo2016synthesized} &  71.2             & 46.6         & 54.1            & 55.6            & 59.1            & 56.3            \\ \hline
SAE~\cite{kodirov2017semantic}        & 80.2             & 54.1            & 33.4            & 33.3            & 42.4            & 40.3            \\ \hline
SSZSL~\cite{guo2017zero}              & -                & -              & 55.8           & -               & -                & -               \\ \hline
GVRZSC~\cite{bucher2017generating}    & -               & -               & 60.1            & -               & -                & -               \\ \hline
GFZSL~\cite{verma2017simple}          & 79.3             & 67.0            & 53.0              & 49.2            & 62.9            & 62.6              \\ \hline
CVAE-ZSL~\cite{mishra2018generative}  & -               &  65.8           & -               & 52.1            & -                & 61.7            \\ \hline
SE-ZSL~\cite{kumar2018generalized}    & \textbf{80.8}             & 69.2            & \textbf{60.3}            & \textbf{59.6}            & 64.5            & \textbf{63.4}            \\ \hline
DCN~\cite{liu2018generalized}         & -                & -            & 55.6            & 56.2            & \textbf{67.4}            & 61.8               \\ \hline
JGM-ZSL~\cite{gao2018joint}           & -                & \textbf{69.5}              & -               & 54.9            & -               & 59.0            \\ \hline
RAS+cGAN~\cite{zhang2019adversarial}  & -                & -            & -               & 52.6            & -               & 61.7               \\ \hline
\textbf{Proposed}                                           & \textbf{81.7}        &\textbf{71.3}             &   \textbf{60.8}              &  \textbf{60.3}              &          \textbf{68.9}                &    \textbf{63.5}                 \\ \hline
\end{tabular}
\end{table}

%%%%%%%%%%%%%%%%%%%%%%%%%%%%
\begin{table}[]
\centering
\caption{Ablative study on three datasets with the PS protocol. The reported values are classification accuracy (\%).}
\label{tb:ablation}
\begin{tabular}{l|l|l|l|}
\cline{2-4}
                                  & AWA2          & SUN           & CUB           \\ \hline
\multicolumn{1}{|l|}{OBTL}        & 65.8          & 56.4          & 54.5          \\ \hline
\multicolumn{1}{|l|}{CL}          & 65.3          & 56.2          & 53.7          \\ \hline
\multicolumn{1}{|l|}{OCD+OBTL}    & 70.9          & 62            & 60.5          \\ \hline
\multicolumn{1}{|l|}{OCD+CL}      & 66.5          & 57.6          & 56.8          \\ \hline
\multicolumn{1}{|l|}{OCD+OBTL+CL} & \textbf{71.3} & \textbf{62.1} & \textbf{60.9} \\ \hline
\end{tabular}
\end{table}

%%%%%%%%%%%%%%%%%%%%%%%%%%%%%%%%%%
\begin{table*}[!t]
\begin{center}
\caption{Average per-class classification accuracy (\%) and harmonic mean accuracy of generalized zero-shot learning when test samples can be from either seen (S) or unseen (U) classes. $\mathbf{A}$: U$\rightarrow$S+U, and $\mathbf{B}$: S$\rightarrow$S+U. ({\textbf{Top two performances are highlighted)}}}
\label{tb:GZSL}
\begin{tabular}{|l|l|l|l|l||l|l|l||l|l|l|}
\hline
\multicolumn{1}{|c|}{\multirow{2}{*}{\textbf{Type}}}& \multicolumn{1}{|c|}{\multirow{2}{*}{\textbf{Method}}}      & \multicolumn{3}{c||}{\textbf{AWA2}}                & \multicolumn{3}{c||}{\textbf{CUB}}                & \multicolumn{3}{c|}{\textbf{SUN}} \\ \cline{3-11} 
& \multicolumn{1}{|c|}{}                                      & \textbf{A} & \textbf{B} & \textbf{H} & \textbf{A} & \textbf{B} & \textbf{H} & \textbf{A}     & \textbf{B}    & \textbf{H}       \\ \hline\hline
\multirow{11}{*}{\rotatebox{90}{Non-Generative Models}} & CONSE~\cite{norouzi2013zero}          & 0.5         & 90.6       & 1.0       & 1.6              & 72.2             & 3.1        & 6.8              & 39.9            & 11.6       \\ \cline{2-11} 
& SSE~\cite{zhang2016learning}          & 8.1         & 82.5       & 14.8          & 8.5              & 46.9             & 14.4       & 2.1              & 36.4             & 4.0    \\ \cline{2-11}
& SJE~\cite{akata2015evaluation}        &  8.0           & 73.9       & 14.4       & 23.5             & 59.2             & 33.6       & 14.7             & 30.5             & 19.8    \\ \cline{2-11}
& ESZSL~\cite{romera2015embarrassingly} & 5.9         & 77.8       & 11.0       & 12.6             & 63.8             & 21.0         & 11.0               & 27.9             & 15.8      \\ \cline{2-11}
& SYNC~\cite{changpinyo2016synthesized} & 10.0          & 90.5       & 18.0       & 11.5             & 70.9             & 19.8       & 7.9              & 43.3             & 13.4      \\ \cline{2-11}
& SAE~\cite{kodirov2017semantic}        & 1.1         & 82.2       & 2.2       & 7.8              & 54.0               & 13.6       & 8.8              & 18.0               & 11.8     \\ \cline{2-11}
& LATEM~\cite{xian2016latent}           & 11.5        & 77.3       & 20.0      & 15.2             & 57.3             & 24.0         & 14.7             & 28.8             & 19.5      \\ \cline{2-11}
& ALE~\cite{akata2013label}             & 14.0          & 81.8       & 23.9      & 23.7             & 62.8             & 34.4       & 21.8             & 33.1             & 26.3    \\ \cline{2-11}
&DCN~\cite{liu2018generalized}         & -           & -          & -       & 28.4             & 60.7             & 38.7       & 25.5             & 37.0               & 30.2       \\ \cline{2-11}
&COSMO+LAGO~\cite{atzmon2019adaptive}  &    52.8           & 80.0          & 63.6    & 44.4             & 57.8             & 50.2       & 44.9             & 37.7             & 41.0       \\  \cline{2-11}
& DEVISE~\cite{frome2013devise}         & 17.1        & 74.7       & 27.8      & 23.8             & 53.0               & 32.8       & 16.9             & 27.4             & 20.9     \\ \hline \hline
\multirow{9}{*}{\rotatebox{90}{Generative Models}} & CVAE-ZSL~\cite{mishra2018generative}  & -           & -          & 51.2    & -                & -                & 34.5       & -                & -                & 26.7    \\ \cline{2-11}
&SE-GZSL~\cite{kumar2018generalized}   & 58.3        & 68.1       & 62.8       & 41.5             & 53.3             & 46.7       & 40.9             & 30.5             & 34.9    \\ \cline{2-11}
&JGM-ZSL~\cite{gao2018joint}           & 56.2        & 71.7       & 63.0       & 42.7             & 45.6             & 44.1       & 44.4             & 30.9             & 36.5      \\ \cline{2-11}
&f-CLSWGAN~\cite{xian2018feature}           &    -           & -          & -    & 43.7            & 57.7             & 49.7       & 42.6             & 36.6             & 39.4       \\ \cline{2-11}
&RAS+cGAN~\cite{zhang2019adversarial}  & -           & -          & -       & 31.5             & 40.2             & 35.3       & 41.2             & 26.7             & 32.4       \\ \cline{2-11}
&CADA-VAE~\cite{schonfeld2019generalized}  &    55.8           & 75.0          & \textbf{63.9}    & 51.6             & 53.5             & {\textbf{52.4}}       & 47.2             & 35.7             & 40.6       \\ \cline{2-11}
&GDAN~\cite{huang2019generative}  & 32.1           & 67.5          & 43.5       & 39.3             & 66.7             & 49.5       & 38.1             & 89.9             & {\textbf{53.4}}       \\ \cline{2-11}

&\textbf{Proposed}                                           & 59.5          &  73.4       &   {\textbf{65.7}}          &    44.8             &       59.9           &   \textbf{51.3}         &  44.8                 &    42.9              &  \textbf{43.8}      \\ \hline
\end{tabular}
\end{center}
\end{table*}

%%%%%%%%%%%%%%%%%%%%%%%%%%%%%%%%%

%%%%%%%%%%%%%%%%%%%%%%%%%%%%%%%%

\subsection{Conventional Zero-Shot Learning (ZSL)}
Table~\ref{tb:zsl} summarizes the results of conventional Zero-Shot Learning. The train split of all three datasets has been used to optimize the proposed framework. For the ZSL problem, synthetic hard samples are generated between unseen classes. The classification accuracies obtained on the PS protocol on AWA2, CUB and SUN databases are $71.3\%$, $60.3\%$, and $63.5\%$, respectively. The proposed framework has improved the state-of-art performance on AWA2, SUN, and CUB databases by $1.8$\%, $0.7$\%, and $0.1$\%, respectively. To estimate whether this difference is significant or not, the McNemar Test~\cite{mcnemar1947note} is used. Keeping a significance threshold of 0.05, or 5\%, we have observed that the null hypothesis is rejected for AWA2 and CUB databases, showcasing that the difference is statistically significant for these two databases. However, for the SUN database, the null hypothesis is not rejected, which implies that the difference between the proposed algorithm from SOTA is insignificant. For the SS protocol, the classification accuracies on AWA2, CUB, and SUN databases are $81.2\%$, $60.8\%$, and $68.4\%$, respectively. In general, across the three databases, the proposed algorithm yields one of the best accuracies compared to several existing approaches. 

\subsection{Ablative Study}
The proposed framework OCD-CVAE has utilized multiple loss functions for improving the performance of ZSL/GZSL. Ablation study is conducted to evaluate the effectiveness of each of the components individually and in combination. Table~\ref{tb:ablation} summarizes the results of five settings thus obtained. It can be observed that OCD+OBTL+CL yields the best results, followed by OCD+OBTL. Also, applying only OBTL and only CL yields poor performance, and it can be attributed to lack of sufficient hard samples for the OBTL and CL loss functions to backpropagate the gradient.     

\subsection{Generalized Zero-Shot Learning (GZSL)}
In GZSL, the testing samples can be from either seen or unseen classes. This is a challenging setting where the train and test classes are not completely disjoint, but the samples of the train and test sets are disjoint. Hence, the possibility of overlapping distribution and hard samples increases in the test set. Most of the ZSL algorithms perform poorly on GZSL. It is our assertion that addressing this GZSL requires learning separability in the embedding space (output of the regressor). The results are reported in terms of average per-class classification accuracies for the protocols $\mathbf{A}$ and $\mathbf{B}$, and the final accuracy is the harmonic mean of accuracies (represented as $\mathbf{H}$), which is computed by $2\times \frac{\mathbf{A}\times \mathbf{B}}{\mathbf{A}+\mathbf{B}}$. 

%To mitigate the poor performance of the algorithms, generating an over-complete distribution of all seen and unseen classes would aid to improve the performance. Once the synthetic data has been approximated, it has been combined with seen distribution and used to train the generator and regressor. While using the BTL and CL loss, the algorithm tries to increase the inter-classes and reduces intra-class distance.  

Table~\ref{tb:GZSL} summarizes the results of existing algorithms on the three databases in GZSL settings. The algorithms are segregated into non-generative and generative models. Among the non-generative models, COSMO+AGO~\cite{atzmon2019adaptive} yields the best performance. While considering all the algorithms, it can be observed that utilizing approximated distribution of unseen class by generative models performs better than non-generative models. The proposed method also utilizes CVAE based generative model. We postulate that generating OCD on the train set and utilizing it to optimize the proposed framework leads the network to better generalize on the test set. 

Between the two protocols A and B, as expected, the results on protocol $\mathbf{B}$ which corresponds to the test set with seen classes are better than the results with unseen test set (protocol $\mathbf{A}$). It is interesting to observe that the proposed framework not necessarily yields the best results for the seen test set, but it is among the top three algorithms on the more challenging unseen test protocol for all three databases. Further, it can be observed from Table~\ref{tb:GZSL} that the proposed framework improves state-of-the-art harmonic mean accuracies $\mathbf{H}$ on the AWA2 dataset by $1.8\%$. The proposed algorithm is among the two best performing algorithms on the SUN and CUB databases. It is worth mentioning that the GZSL is a challenging problem, and none of the algorithms has consistently outperformed on all three databases.

\subsection{Hyper-Parameter Selection}

\begin{figure}[!t]
\centering
  \includegraphics[width=0.5\textwidth]{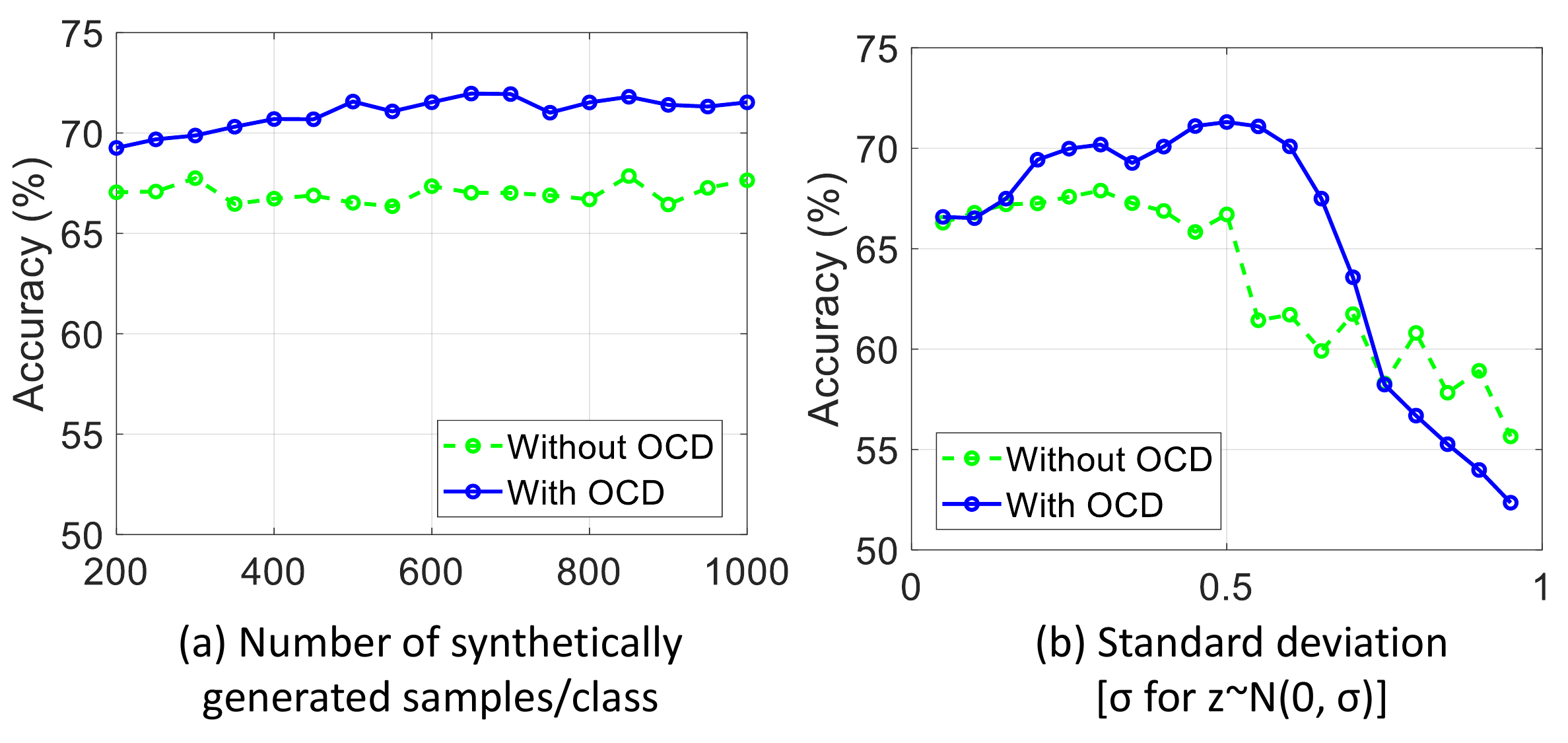}
  \caption{Hyper-parameter selection on the AWA2 dataset with the PS protocol. Accuracy plots by varying (a) the number of samples and (b) standard deviation.}
  \label{fig:stds}
\end{figure}

Figure~\ref{fig:stds}(a) shows the performance with an increasing number of synthetically generated samples. It can be observed that when OCD is not used for training the regressor, increasing the number of samples does not affect the performance. With the use of OCD, generating $400$ to $600$ samples leads to improved performance. To determine the value of $\sigma^{'}_{HP}$ in Equation~\ref{eq:ocd2}, we have explored the range of $\sigma^{'}_{HP}$ from $0.05$, to $0.95$. As shown in Figure~\ref{fig:stds}(b), it can be observed that the best performance has been achieved on $0.5$ standard deviation. The value of $\sigma_{HP}$ is chosen from a standard normal, while $\sigma_{HP}’$ is computed using the PS split train set. The results in Figure~\ref{fig:stds} also demonstrate that the chosen value ($0.5$) yields the best results. Most of the hyper-parameters are kept consistent with Verma~\etal~\cite{kumar2018generalized}. In OBTL loss, $\alpha$ parameter is computed during optimization on the train set and is set as $0.4$. 

\section{Conclusion}

This paper addresses the challenge of Zero-Shot Learning and Generalized Zero-Shot Learning. We propose the concept of over-complete distribution and utilize it to train the discriminative classifier in ZSL and GZSL settings. An over-complete distribution is defined by generating all possible hard samples for a class which are closer to other competing classes. We have observed that over-complete distributions are helpful in ensuring separability between classes and improve the classification performance. Experiments on three benchmark databases with both ZSL and GZSL protocols show that the proposed approach yields improved performance. The concept of OCD along with optimizing inter-class and intra-class distances can also be utilized in other frameworks such as Generative Adversarial Networks, heterogeneous metric learning \cite{sdg}, and applications such as face recognition with disguise variations \cite{dfw}. 

%Experimentally, we have shown that learning a classifier on separable distribution would not generalize on test samples with the class-overlapping distributions. 

\section{Acknowledgement}
%This publication is an outcome of the R\&D work undertaken a project under the Visvesvaraya PhD Scheme of Ministry of Electronics \& Information Technology, Government of India, being implemented by Digital India Corporation. 
R. Keshari is partially supported by Visvesvaraya Ph.D. Fellowship. M. Vatsa is partly supported by the Swarnajayanti Fellowship, Government of India.

{\footnotesize
\bibliographystyle{ieee_fullname}
\bibliography{egbib}

\begin{thebibliography}{10}\itemsep=-1pt

\bibitem{abadi2015tensorflow}
Mart{\i}n Abadi, Ashish Agarwal, Paul Barham, Eugene Brevdo, Zhifeng Chen,
  Craig Citro, Greg~S Corrado, Andy Davis, Jeffrey Dean, Matthieu Devin, et~al.
\newblock Tensorflow: Large-scale machine learning on heterogeneous systems,
  2015.
\newblock {\em Software available from tensorflow. org}, 1(2), 2015.

\bibitem{akata2013label}
Zeynep Akata, Florent Perronnin, Zaid Harchaoui, and Cordelia Schmid.
\newblock Label-embedding for attribute-based classification.
\newblock In {\em CVPR}, pages 819--826, 2013.

\bibitem{akata2015evaluation}
Zeynep Akata, Scott Reed, Daniel Walter, Honglak Lee, and Bernt Schiele.
\newblock Evaluation of output embeddings for fine-grained image
  classification.
\newblock In {\em CVPR}, pages 2927--2936, 2015.

\bibitem{Amos2016OpenFace}
Brandon Amos.
\newblock Openface 0.2.0: Higher accuracy and halved execution time.
\newblock \url{http://bamos.github.io/2016/01/19/openface-0.2.0/}, 2016.

\bibitem{atzmon2019adaptive}
Yuval Atzmon and Gal Chechik.
\newblock Adaptive confidence smoothing for generalized zero-shot learning.
\newblock In {\em CVPR}, pages 11671--11680, 2019.

\bibitem{bucher2017generating}
Maxime Bucher, St{\'e}phane Herbin, and Fr{\'e}d{\'e}ric Jurie.
\newblock Generating visual representations for zero-shot classification.
\newblock In {\em ICCV}, pages 2666--2673, 2017.

\bibitem{caruana1997multitask}
Rich Caruana.
\newblock Multitask learning.
\newblock {\em Machine learning}, 28(1):41--75, 1997.

\bibitem{changpinyo2016synthesized}
Soravit Changpinyo, Wei-Lun Chao, Boqing Gong, and Fei Sha.
\newblock Synthesized classifiers for zero-shot learning.
\newblock In {\em CVPR}, pages 5327--5336, 2016.

\bibitem{frome2013devise}
Andrea Frome, Greg~S Corrado, Jon Shlens, Samy Bengio, Jeff Dean, Tomas
  Mikolov, et~al.
\newblock Devise: A deep visual-semantic embedding model.
\newblock In {\em NIPS}, pages 2121--2129, 2013.

\bibitem{gao2018joint}
Rui Gao, Xingsong Hou, Jie Qin, Li Liu, Fan Zhu, and Zhao Zhang.
\newblock A joint generative model for zero-shot learning.
\newblock In {\em ECCV}, pages 631--646. Springer, 2018.

\bibitem{sdg}
Soumyadeep Ghosh, Mayank Vatsa, and Richa Singh.
\newblock Subclass heterogeneity aware loss for cross-spectral cross-resolution
  face recognition.
\newblock {\em IEEE Transactions on Biometrics, Behavior, and Identity
  Science}, 2020.

\bibitem{guo2017zero}
Yuchen Guo, Guiguang Ding, Jungong Han, and Yue Gao.
\newblock Zero-shot learning with transferred samples.
\newblock {\em TIP}, 26(7):3277--3290, 2017.

\bibitem{hu2017toward}
Zhiting Hu, Zichao Yang, Xiaodan Liang, Ruslan Salakhutdinov, and Eric~P Xing.
\newblock Toward controlled generation of text.
\newblock In {\em ICML}, pages 1587--1596. JMLR. org, 2017.

\bibitem{huang2019generative}
He Huang, Changhu Wang, Philip~S Yu, and Chang-Dong Wang.
\newblock Generative dual adversarial network for generalized zero-shot
  learning.
\newblock In {\em CVPR}, pages 801--810, 2019.

\bibitem{kodirov2017semantic}
Elyor Kodirov, Tao Xiang, and Shaogang Gong.
\newblock Semantic autoencoder for zero-shot learning.
\newblock In {\em CVPR}, pages 3174--3183, 2017.

\bibitem{lampert2009learning}
Christoph Lampert, Hannes Nickisch, and Stefan Harmeling.
\newblock Learning to detect unseen object classes by between-class attribute
  transfer.
\newblock In {\em CVPR}, pages 951--958. IEEE, 2009.

\bibitem{lampert2014attribute}
Christoph~H Lampert, Hannes Nickisch, and Stefan Harmeling.
\newblock Attribute-based classification for zero-shot visual object
  categorization.
\newblock {\em TPAMI}, 36(3):453--465, 2014.

\bibitem{larochelle2008zero}
Hugo Larochelle, Dumitru Erhan, and Yoshua Bengio.
\newblock Zero-data learning of new tasks.
\newblock In {\em AAAI}, volume~1, page~3, 2008.

\bibitem{liu2018generalized}
Shichen Liu, Mingsheng Long, Jianmin Wang, and Michael~I Jordan.
\newblock Generalized zero-shot learning with deep calibration network.
\newblock In {\em NIPS}, pages 2009--2019, 2018.

\bibitem{long2018zero}
Yang Long, Li Liu, Fumin Shen, Ling Shao, and Xuelong Li.
\newblock Zero-shot learning using synthesised unseen visual data with
  diffusion regularisation.
\newblock {\em TPAMI}, 40(10):2498--2512, 2018.

\bibitem{mcnemar1947note}
Quinn McNemar.
\newblock Note on the sampling error of the difference between correlated
  proportions or percentages.
\newblock {\em Psychometrika}, 12(2):153--157, 1947.

\bibitem{miller2002learning}
Erik~Gundersen Miller.
\newblock {\em Learning from one example in machine vision by sharing
  probability densities}.
\newblock PhD thesis, Massachusetts Institute of Technology, 2002.

\bibitem{mishra2018generative}
Ashish Mishra, Shiva Krishna~Reddy, Anurag Mittal, and Hema~A Murthy.
\newblock A generative model for zero shot learning using conditional
  variational autoencoders.
\newblock In {\em CVPRW}, pages 2188--2196, 2018.

\bibitem{norouzi2013zero}
Mohammad Norouzi, Tomas Mikolov, Samy Bengio, Yoram Singer, Jonathon Shlens,
  Andrea Frome, Greg~S Corrado, and Jeffrey Dean.
\newblock Zero-shot learning by convex combination of semantic embeddings.
\newblock {\em arXiv preprint arXiv:1312.5650}, 2013.

\bibitem{patterson2012sun}
Genevieve Patterson and James Hays.
\newblock Sun attribute database: Discovering, annotating, and recognizing
  scene attributes.
\newblock In {\em CVPR}, pages 2751--2758. IEEE, 2012.

\bibitem{raina2007self}
Rajat Raina, Alexis Battle, Honglak Lee, Benjamin Packer, and Andrew~Y Ng.
\newblock Self-taught learning: transfer learning from unlabeled data.
\newblock In {\em ICML}, pages 759--766. ACM, 2007.

\bibitem{richardson1997bayesian}
Sylvia Richardson and Peter~J Green.
\newblock On bayesian analysis of mixtures with an unknown number of components
  (with discussion).
\newblock {\em RSS: series B (statistical methodology)}, 59(4):731--792, 1997.

\bibitem{romera2015embarrassingly}
Bernardino Romera-Paredes and Philip Torr.
\newblock An embarrassingly simple approach to zero-shot learning.
\newblock In {\em ICML}, pages 2152--2161, 2015.

\bibitem{schonfeld2019generalized}
Edgar Schonfeld, Sayna Ebrahimi, Samarth Sinha, Trevor Darrell, and Zeynep
  Akata.
\newblock Generalized zero-and few-shot learning via aligned variational
  autoencoders.
\newblock In {\em CVPR}, pages 8247--8255, 2019.

\bibitem{dfw}
Maneet {Singh}, Richa {Singh}, Mayank {Vatsa}, Nalini~K. {Ratha}, and Rama
  {Chellappa}.
\newblock Recognizing disguised faces in the wild.
\newblock {\em IEEE Transactions on Biometrics, Behavior, and Identity
  Science}, 1(2):97--108, 2019.

\bibitem{thrun1996learning}
Sebastian Thrun.
\newblock Is learning the n-th thing any easier than learning the first?
\newblock In {\em NIPS}, pages 640--646, 1996.

\bibitem{kumar2018generalized}
Vinay~Kumar Verma, Gundeep Arora, Ashish Mishra, and Piyush Rai.
\newblock Generalized zero-shot learning via synthesized examples.
\newblock In {\em CVPR}, pages 4281--4289, 2018.

\bibitem{verma2017simple}
Vinay~Kumar Verma and Piyush Rai.
\newblock A simple exponential family framework for zero-shot learning.
\newblock In {\em ECML}, pages 792--808. Springer, 2017.

\bibitem{welinder2010caltech}
Peter Welinder, Steve Branson, Takeshi Mita, Catherine Wah, Florian Schroff,
  Serge Belongie, and Pietro Perona.
\newblock Caltech-ucsd birds 200.
\newblock 2010.

\bibitem{wen2016discriminative}
Yandong Wen, Kaipeng Zhang, Zhifeng Li, and Yu Qiao.
\newblock A discriminative feature learning approach for deep face recognition.
\newblock In {\em ECCV}, pages 499--515. Springer, 2016.

\bibitem{xian2016latent}
Yongqin Xian, Zeynep Akata, Gaurav Sharma, Quynh Nguyen, Matthias Hein, and
  Bernt Schiele.
\newblock Latent embeddings for zero-shot classification.
\newblock In {\em CVPR}, pages 69--77, 2016.

\bibitem{xian2018zero}
Yongqin Xian, Christoph~H Lampert, Bernt Schiele, and Zeynep Akata.
\newblock Zero-shot learning-a comprehensive evaluation of the good, the bad
  and the ugly.
\newblock {\em TPAMI}, 2018.

\bibitem{xian2018feature}
Yongqin Xian, Tobias Lorenz, Bernt Schiele, and Zeynep Akata.
\newblock Feature generating networks for zero-shot learning.
\newblock In {\em CVPR}, pages 5542--5551, 2018.

\bibitem{xian2017zero}
Yongqin Xian, Bernt Schiele, and Zeynep Akata.
\newblock Zero-shot learning-the good, the bad and the ugly.
\newblock In {\em CVPR}, pages 4582--4591, 2017.

\bibitem{zhang2019triple}
Haofeng Zhang, Yang Long, Yu Guan, and Ling Shao.
\newblock Triple verification network for generalized zero-shot learning.
\newblock {\em TIP}, 28(1):506--517, 2019.

\bibitem{zhang2019adversarial}
Haofeng Zhang, Yang Long, Li Liu, and Ling Shao.
\newblock Adversarial unseen visual feature synthesis for zero-shot learning.
\newblock {\em Neurocomputing}, 329:12--20, 2019.

\bibitem{zhang2016learning}
Ziming Zhang and Venkatesh Saligrama.
\newblock Learning joint feature adaptation for zero-shot recognition.
\newblock {\em arXiv preprint arXiv:1611.07593}, 2016.

\end{thebibliography}
}

\end{document}